\documentclass[11pt]{article}
\newcommand{\ra}{\mbox{$\rightarrow$}}
\newcommand{\rt}{\mbox{$\rightarrow$~}}

\newcommand{\m}[1]{{\mbox{\tt #1}}}           

\newcommand{\nil}[1]{}
\newenvironment{ite}{                     
     \parskip 0cm \begin{itemize} \parskip 0cm \parsep 0cm \itemsep 0cm \topsep 0cm}{
        \end{itemize}} 

\usepackage{graphicx}
\begin{document}
\title{Story Generation and Aviation Incident Representation: \\
Working Note 14}
\author{Peter Clark \\
Knowledge Systems \\
Applied Research and Technology \\
The Boeing Company \\
peter.e.clark@boeing.com}
\date{Jan 26th 1999}

\maketitle

\bibliographystyle{apalike}                   

\begin{abstract}
This working note discusses the topic of story generation, with
a view to identifying the knowledge required to understand
aviation incident narratives (which have structural similarities
to stories), following the premise that to understand aviation
incidents, one should at least be able to generate examples of
them. We give a brief overview of aviation incidents and their
relation to stories, and then describe two of our earlier attempts
(using `scripts' and `story grammars') at incident generation 
which did not evolve promisingly. 
Following this, we describe a simple incident generator which
did work (at a `toy' level), using a `world simulation'
approach. This generator is based on Meehan's TALE-SPIN story
generator \cite{tale-spin}. 
We conclude with a critique of the approach.
\end{abstract}

\section{Introduction}

This working note discusses the topic of story generation, with
a view to identifying the knowledge required to understand
aviation incident narratives (which have structural similarities
to stories), following the premise that to understand aviation
incidents, one should at least be able to generate examples of
them. We give a brief overview of aviation incidents and their
relation to stories, and then describe two of our earlier attempts
(using `scripts' and `story grammars') at incident generation 
which did not evolve promisingly. 
Following this, we describe a simple incident generator which
did work (at a `toy' level), using a `world simulation'
approach. This generator is based on Meehan's TALE-SPIN story
generator \cite{tale-spin}. 
We conclude with a critique of the approach.

Aviation incident reports describe and analyse unusual/unexpected 
chains of events during aircraft operation (eg. mechanical
problems, unusual maneuvers). The part of the 
report describing the incident itself is typically an episodic
narrative, and has similarities in structure to {\it stories}, 
already studied in the linguistics literature. As part of a speculative 
exploration of achieving computer-based understanding of aviation incidents
(that is, having a computer be able to answer various questions about
incidents, including facts not explicitly stated in the narrative),
we discuss and illustrate automatic generation of `toy' 
aviation-incident-like reports.
The premise here is that, if a computer is to be able
to understand aviation incidents, it should at least be able to
generate plausible incident `stories'. This requirement is 
quite a strong one, but is imposed here to ensure that 
key information about aviation incidents is captured; or, viewed 
another way, this requirement provides a good guide for delineating
what a knowledge base about aviation safety should contain. 
It is also imposed to `keep the knowledge engineer honest':
`understanding-only' systems can sometimes give the impression
of understanding more than they really do\footnote{For example,
anecdotally one of the rules used for interpreting terrorism news articles
in the Message Understanding Conference (MUC) Competitions was:
``if you see an integer which is a multiple of ten, then this is 
probably the number of victims in the terrorist incident.''}.

Some examples of very brief aviation incident summaries are given 
below, from the FAA's Incident Data Systsem (FIDES) publically available
at \verb+http://nasdac.faa.gov/+ \verb+asp/fw_fids.asp+:

\small \begin{verbatim}
   961222042819C
   INOPERATIVE TRANSPONDER ON CLIMBOUT FROM AIRPORT. RETURNED. PUT IT ON 
      MEL. CONTINUED FLIGHT. 
 
   961211044319C
   PASSENGER CUSSED OUT FLIGHT ATTENDANT TAXIING TO RUNWAY. PIC RETURNED 
      TO GATE. PASSENGER REMOVED.   

   961216043479C
   TURBINE RIGHT ENGINE FAILED. DIVERTED TO RIC. OVERWEIGHT LANDING. HAD 
      CONTAINED TURBINE FAILURE. 

   960712045359C
   PIC BECAME INCAPACITATED AFTER LANDING. LANDING WAS ERRATIC AS WAS 
      TAXIING. FIRST OFFICER TOOK OVER. PIC HAD STROKE
\end{verbatim} \normalsize

In fact, aviation incidents are not quite the same as `stories':
although they contain plausible sequences of
events, the sequences are non-fictional, and have not been crafted to
deliberately interesting or suspenseful (though of course many are);
similarly the narrative is intended to be clear and factual,
rather than written with suspense and intrigue. These 
differences are important when considering the relevance
of story generation systems to incident generation.

Early work on story generation systems
(most notably Meehan's TALE-SPIN \cite{tale-spin})
treated story generation as synonymous with world
simulation, where the world contained (possibly several)
agents who rationally tried to achieve their goals.
This task in itself is an interesting and challenging
one, and characterizes well what  an aviation incident 
generation system should do. Later work, eg. \cite{dehn}, pointed 
out that story generation is more than just simulation: the
author him/herself has goals (eg. create an interesting
story), and will explore multiple ways of unfolding a
story to best achieve those goals. A story generator
thus involves not only world simulation, but also
meta-level control manipulating both the way that
simulation plays out, and the characters/places/world 
being simulated. In addition, as Smith and Witten point
out \cite{tailor}, there is an important distinction between 
the story itself and the telling of it (eg. events may
be related in non-chronological order), a distinction which
almost all computer-based story generators fail to make.
However, for our goals of aviation incident generation, 
a simulation-only approach is a good starting point, and one 
which we focus on in this paper.

\section{Approaches to Episodic Narrative Generation}

Before describing this `simulation' approach, we first
describe two other approaches which we explored, both
based on using pre-defined fragments of event sequences
which might occur. Neither of these
seemed to be evolving into a practical system, so they were
abandoned. However, it is worth briefly mentioning them.
It is also worth noting that the `scripts' and `story
grammars' were originally proposed in the literature for
story understanding, not generation, so our difficulties
do not necessarily reflect a problem with these ideas, but
with an attempt to make them do something that they were
originally not designed for (but might plausibly have
been useful for). In Section~\ref{story-grammars} we return
to offer further reflections on story grammars with
hindsight.

\subsection{A Script-based Approach}

\begin{figure}
\includegraphics[width=150mm]{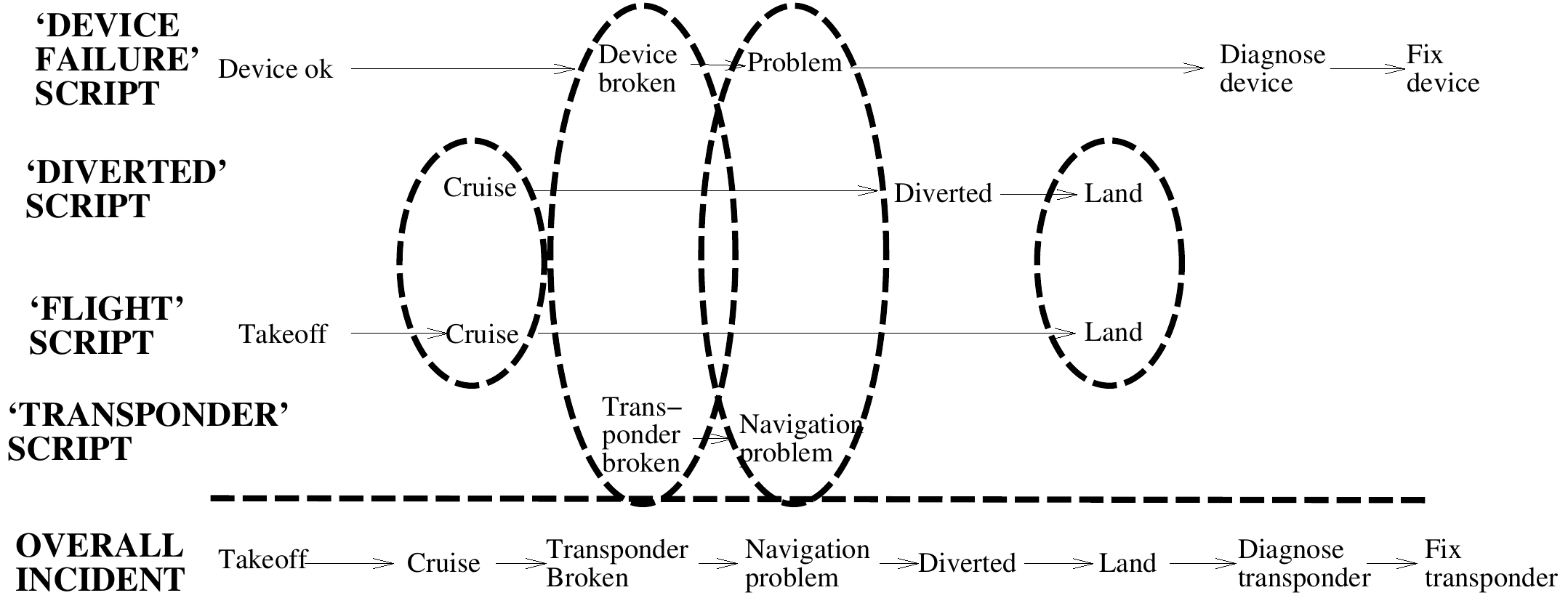}
\caption{Component scripts combined into an overall incident scenario. 
The ovals denote coreferential events in multiple scripts. The four
scripts are combined to produce the overall incident shown on the
last line.
\label{scripts}}
\end{figure}

In a script-based approach, a library of scripts is manually created,
describing different abstract sequences of events appropriate to aviation 
incidents. To generate an incident, the computer selects and
appropriately interleaves a suitable set of scripts.
For example, consider the component scripts, and their
interleaving to create an incident, shown in Figure~\ref{scripts}.

Several challenges became apparent with this approach. First, 
there are a lot of scripts, including many simple variants of
the same general theme.
For example, the conceptual idea of `aborted flight' has
many associated scripts (``load\ra unload'', 
``load\ra taxi-out\ra taxi-out\ra taxi-back\ra unload'', 
``load\ra taxi-out\ra takeoff\ra return\ra taxi-back\ra unload'', ...). 
Second, there are many causal interactions and constraints
between scripts, making the proper and automatic
interleaving of scripts extremely complicated (eg.
the plane should be diverted {\it after} a problem
has occurred, not before, and only if it's a serious
problem). Similarly, the choice of which scripts are
appropriate and combinable is difficult to make. Finally,
the scripts contain no representation of {\it why} an
event occurs (eg. the plane returns to the start airport
{\it because} it is important to land as soon as possible,
where it is safe), making this approach intuitively less
appealing.

\subsection{A Story Grammar Approach \label{storygrammars1}}

Several authors have proposed the use of `story grammars'
to capture the structure of stories (eg. \cite{rumelhart}, and
see \cite{story-grammars} for a discussion). For example, an `aviation incident' grammar
might be as follows, using Definite Clause Grammer notation:

\small \begin{verbatim}
   incident --> start-of-flight, problem(P), response(P).

   start-of-flight --> [taxi].
   start-of-flight --> [taxi,takeoff].
   start-of-flight --> [taxi,takeoff,cruise].

   problem(broken(P)) --> broken(P).
   problem(bad_weather(W)) --> bad_weather(W).

   broken(transponder) --> [transponder_broke].

   bad_weather(stormy) --> [stormy].

   response(broken(transponder)) --> return_to_ground.

   ...
\end{verbatim} \normalsize

Again, this approach presents some challenges. First, as with
scripts, there are often many situation-specific ways of realizing
a single conceptual action. For example, the notion of 
\verb+return_to_ground+ above may involve aborted takeoff, 
divert to nearest airport, continue with landing, or no action at 
all (if the airplane is already on the ground). The grammar starts
to look messy when attempting to account for these different options.
Second, an incident often involves interleaved events, for example:
The plane took off, then a passenger was ill, then bad weather hit, ...
Stories (at least aviation incidents) appear to be less constrained 
and more multithreaded than the structure of an English sentence, 
making grammars less suitable for modeling their structure.

\section{A World Simulation Approach \label{mysystem}}

We now describe a simple incident generation system,
which is essentially a simplified reconstruction of
Meehan's TALE-SPIN system \cite{tale-spin}. It also has some
interesting similarities to Smith and Witten's TAILOR system
\cite{tailor}, which we summarize later in Section~\ref{tailor}. 

TALE-SPIN generates simple fables, using a world simulator in
which events unfold. Rather than being based on pre-defined 
fragments of sequences, rational sequences of actions are created 
on-the-fly using a backward-chaining planning algorithm. An example story 
from TALE-SPIN is:
\begin{quote}
``Once upon a time, there was a dishonest fox named Henry
who lived in a cave, and a vain and trusting crow named Joe
who lived in an elm tree. Joe had gotten a piece of cheese and
was holding it in his mouth. One day, Henry walked from his
cave, across the medow to the elm tree. He saw Joe Crow
and the cheese and became hungry. He decided that he might
get the cheese if Joe Crow spoke, so he told Joe that he liked
his singing very much and wanted to hear him sing. Joe
was very pleased with Henry and began to sing. The cheese fell out of
his mouth, down to the ground. Henry picked up the
cheese and told Joe Crow that he was stupid. Joe was angry, and didn't
trust Henry anymore. Henry returned to his cave.'' \cite[p97]{tale-spin}
\end{quote}

In some ways, TALE-SPIN can be viewed as a generalization of a 
STRIPS-like planner\footnote{ie. backward-chaining from a goal,
using a precondition/effects
action representation, although TALE-SPIN's actual implementation appears
cruder than STRIPS (eg. does it maintain a ``protected goal'' list?)}
 \cite{strips}. The original STRIPS planner 
produced a sequence of actions to achieve a pre-defined 
goal, that is it produced a `rational' plan. However, 
it operated in an inert world -- the actions were 
deterministic and guaranteed to work, no surprises 
or problems occured (eg. the tower falls down when 
stacking blocks). In contrast, TALE-SPIN's world is more 
dynamic: there are multiple characters pursuing different goals,
and (presumably) non-volitional `happenings' can also occur
in the world, requiring the characters to react.

Our `reconstruction' is simple and highly improvised, based
only on the general ideas in Meehan's 1977 paper (rather than 
a detailed reimplementation of his system). The reconstruction
undoubtedly misses out a lot of Meehan's program, and also adds in
new parts and approaches that Meehan didn't originally conceive of 
or emphasize. The source Prolog code for the `reconstruction' is 
given in the Appendix (written from scratch for this exercise).
The incidents generated are fictitious, toy examples -- the goal 
in this brief exercise was to investigate the structure
of a generation system, rather than aim for detailed
fidelity to true incidents. The reader should be aware that
the simplicity of these toy incidents is in stark contrast
to the complexity of real aviation incidents.

In our reconstruction, an incident is represented as a sequence
of events, which transforms an initial situation to a final situation.
A situation is represented as a set of facts, for example
the start situation in all the incidents generated is:
\begin{center}
\small \begin{tabular}{ll}
\verb+[ plocation(passengers1, gate(seattle)),+ & {\it \% the passengers are at the Seattle gate} \\
\verb+  alocation(airplane1, gate(seattle)), + & {\it \% the airplane is at the Seattle gate} \\
\verb+  flight_path(seattle, chicago),       + & {\it \% can fly from Seattle to Chicago} \\
\verb+  flight_path(chicago, dallas),        + & {\it \% can fly from Chicago to Dallas} \\
\verb+  airplane(airplane1),                 + & {\it \% airplane1 is an airplane} \\
\verb+  passengers(passengers1) ],           + & {\it \% passengers1 is a group of passengers} \\
\end{tabular} \normalsize
\end{center}

An event is represented in the usual STRIPS fashion,
with precondition facts, an add-list of facts which become true
after the event has occured, and a delete-list of facts which
become false. For example, taxiing to the runway is represented:
\begin{center}
\small \begin{tabular}{ll}
\verb+ed(action, take_off(Airplane,Airport),		 + & {\it \% To take off...} \\
\verb+   /*pcs*/ [alocation(Airplane,runway(Airport))],+ & {\it \% must be on the runway...} \\
\verb+   /*del*/ [alocation(Airplane,runway(Airport))],+ & {\it \% result: no longer on runway..} \\
\verb+   /*add*/ [alocation(Airplane,near(Airport))],  + & {\it \% ..but (just) near the airport} \\
\verb+   /*txt*/ ['The plane took off from ',Airport,'.']).+ & \\ 
\end{tabular} \normalsize \end{center}

An event is either an {\it action}, which the pilot can perform (eg. \m{take\_off}, above),
or a {\it happening}, which is out of his/her control (eg. \m{ill\_passenger}, meaning
a passenger becomes ill).
A STRIPS-like planner forms
one component, and is used to generate a plan (action sequence) for the airplane 
pilot to achieve his/her goal (initially, to get the passengers
to Dallas, ie. to achieve \m{plocation(passengers1, gate(dallas))}).
There is also an independent `simulator' component
which simulates execution of that plan. Given a plan, the simulator
will execute it one step at a time. However, it may also (with some
probability) make a `happening' occur
during the execution, causing
the current situation to change unexpectedly. When a
happening occurs, the original goal (of the pilot) is
reassessed -- for example, the original goal of `get the passengers
to Dallas' (\m{plocation(passengers1, gate(dallas))}) 
may change to `get the passengers
to safety as quickly as possible' (\m{p\_on\_ground(passengers1)}).
If the goal changes, then
the pilot's current plan is abandoned, and a new plan is generated on
the fly to achieve the new goal from the current situation. 
The simulator then continues
with the simulation, executing the new plan. In this
implementation, just one happening is introduced in an incident,
but in principle several could occur. The incident ends
when the pilot achieves his/her (current) goal.
English text is generated from the incident representation
using simple fill-in-the-blank text templates, and was
not a focus of this short project.
\noindent
An example (annotated) incident generated is the following:
\begin{quote}
\noindent
{\it Initially, the pilot goal = get the passengers to Dallas.
Hence initial plan = taxi \ra take-off \ra cruise to Dallas \ra land \ra taxi to gate.
The simulator starts executing the plan...} \\
\verb+        The passengers boarded the plane.+ \\
\verb+        The plane taxiied to the runway.+ \\
\verb+        The plane took off from seattle.+ \\
{\it Now a happening is introduced...} \\
\verb+        A passenger became very ill.+ \\
{\it The pilot revises his/her goal, to be get medical help for the passenger.
Hence new plan = land \ra taxi to gate \ra get medical help.} \\
\verb+        The plane landed at seattle.+ \\
\verb+        The plane taxiied to the gate.+ \\
\verb+        The passengers disembarked.+ \\
\verb+        Medical help was provided.+
\end{quote}

Generation of an incident thus involves the rational 
pursuit of goals, combined with the introduction of
unexpected events (`happenings') and rational modification
of goals in the light of the new facts. Some other 
example incidents generated by the system are shown below:

\small \begin{verbatim}
              The passengers boarded the plane.
              The plane taxiied to the runway.
              The plane took off from seattle
              The plane cruised towards chicago
              The plane cruised towards dallas
              The engine caught fire.
              The plane landed at dallas
              The passengers were evacuated from the plane.
              ----------
              The passengers boarded the plane.
              A passenger became very ill.
              The passengers disembarked.
              Medical help was provided.
              ----------
              The passengers boarded the plane.
              The plane taxiied to the runway.
              The plane took off from seattle
              The plane cruised towards chicago
              The engine caught fire.
              The plane landed at chicago
              The passengers were evacuated from the plane.
\end{verbatim} \normalsize

\noindent
The knowledge required to generate these is as follows:
\begin{enumerate}
\item A STRIPS-like representation of the actions the
              pilot can perform (takeoff, land, etc.).
        These describe the preconditions for the action,
        the facts which become false as a result of performing
        the action, and the facts which become true.
\item A similar STRIPS-like representation of `happenings' which
              can occur.
\item A representation of how the pilot goals should change
              should something unexpected occur (eg. IF ill patient 
              THEN new goal is to get medical help).
\item A scoring function, for ranking different plans and choosing the best.
\end{enumerate}
This last item is particularly interesting -- a `rational'
plan is not only one which achieves the goal, but one
which is of minimum cost. For example, the following
`irrational' incidents were generated during debugging
the system, when plans were not ranked according to
their quality:

\small \begin{verbatim}
              The passengers boarded the plane.
              The plane taxiied to the runway.
              The engine caught fire.
              The plane took off from seattle.
              The plane cruised towards chicago.
              The plane cruised towards dallas.
              The pilot made an emergency landing near somewhere.
              The passengers were evacuated from the plane.
\end{verbatim} \normalsize
In this case, the engine fire causes a new goal of 
`get the plane to the ground'. However, the plan of
taking off and landing again is considered just as
effective as staying put, and (here) is carried out.

Similarly in this buggy incident:

\small \begin{verbatim}
              The passengers boarded the plane.
              The plane taxiied to the runway.
              The plane took off from seattle
              The plane cruised towards chicago.
              The plane cruised towards dallas.
              The plane landed at dallas.
              The passengers were evacuated from the plane.
\end{verbatim} \normalsize
evacuating the plane (even though there is no problem) is 
considered just as good a way of getting the pasengers to 
Dallas as delivering them right to the gate.

\section{Discussion}

This simple TALE-SPIN-like system is able to generate
plausible `toy' aviation incidents; part of the reason it
is able to do this is that the underlying representation
captures the relationships between goals, plans, actions, 
and happenings. It could thus, in principle, answer questions such as
``Why did the pilot do X?'', or ``Why did X happen?''.
Representing these relationships seems essential to
the computer having some reasonably deep model of
what is happening during an incident.

There are several simplifications in this small
implementation which are worth noting. None seem
to be fundamental obstacles to making the generator
richer, and we speculate on how the system could be
generalized:
\begin{enumerate}
\item {\bf Deterministic actions}. In the implementation,
actions are deterministic, ie. their outcomes are
certain. To account for non-deterministic actions:
\begin{ite}
\item the STRIPS-like action representation would have to be 
              generalized to represent the possible outcomes (eg.
	as different sets of add/delete lists), and their
              associated probabilities.
\item the planner would need to to explore possible
              branches of a plan's execution, and their probability
              of occurence. The overall `quality' of a plan would be
              some function of its possible outcomes and
              their associated probability.
\item the simulator would similarly need to randomly select
              one possible outcome of an action, again based on
              the outcomes' associated probabilities.
\end{ite}
\item {\bf Linear plans.} In this implementation, a 
plan is a sequence of actions. However, a more sophisticated
approach would allow for conditional plans (eg. ``do X,
then if Y happens do Z$_{1}$ else do Z$_{2}$''), conceptually
more like a flow-chart than a list of actions. 
This would make the planner
more complex, but would allow it to account for different
potential outcomes of an action, and even anticipate possible
happenings occuring and plan corrective actions for them ahead 
of time\footnote{This forms an important part of (real) pilot
training, where, for example, pilots should be constantly
aware of where they plan to land if there is an emergency.}.
Here, the planning algorithm starts to ressemble game-playing 
algorithms, where a branching tree of actions and possible
reactions is searched to decide the best way to proceed.
If we assume the pilot has a good model of the world,
the planning algorithm should use the simulation algorithm
to develop a plan (in the current implementation, the
planning algorithm uses a different and simpler `simulation'
engine, in which happenings do not occur).
\item {\bf Multiple agents.} In the system, there
is just one agent (the pilot), interacting with a world.
A more sophisticated approach would allow for multiple
agents, each of whom would be pursuing his/her own plan to
achieve his/her own goals, and the interaction of those
pursuits causing interesting events to occur. TALE-SPIN
appears to maintain plans for more than one agent in
its simulator. Similarly, the implementation could be
extended in this way (other agents might include 
air traffic control, the passengers, the flight attendants).
\item {\bf Goal revision.} At present, the implementation
contains shallow rules for revising a goal in the
light of new facts. For example, if there is an 
engine fire, the goal is changed to `land as quickly
as possible' but there is no representation of {\it why}
that is an appropriate new goal. To represent this
in a deeper way (and hence tolerate genuinely unexpected 
happenings), the system would need to explore and
compare consequences of doing nothing versus taking various
alternative corrective actions. This would need to
be a directed search, ie. not simply explore all
possible random action sequences in response to an
unexpected event, but instead identify what the
undesirable outcome was, and identify actions which
prevent/reduce the probability of that undesirable
outcome. A concept like `minimize the time in the air'
would require some thought as to how it could be
added to the representation.
\item {\bf First principles planning.} All the plans in
the implementation are generated `from first principles', 
using the STRIPS-like planner. While this allows the 
system to simulate the pilot's response to many different 
events, it sometimes seems overkill and counterintuitive: 
for example, in the simulation the pilot works out from first principles
how to get the passengers to Dallas (he/she needs to land there,
thus he/she must fly there, thus he/she must take off, ...),
rather than just simply follow a predefined flight plan.
Although scripts were abandoned earlier as the {\it sole}
means of generating incident scenarios, it seems that
there should at least be a place for them in the 
incident generator (for example, a script might encode
the official procedure for response to a particular
happening, which the pilot decides to follow). Again,
there does not seem to be any major obstacles to including
this.
\end{enumerate}

\section{Some Related Work Revisited}

\subsection{Story Grammars \label{story-grammars}}

In 1975, Rumelhart published a compelling and influential 
article, suggesting that {\it story grammars} would account 
for the structure of stories \cite{rumelhart}. We offer
some additional reflections on his paper here, in particular
why a grammar does not appear adequate for story
{\it generation}.

Similar to our incident representation, Rumelhart essentially
suggests that a story is a hierarchical set of events,
with goal-based constaints between them. (Slightly modified) examples 
of his grammar rules include:

\small \begin{center} \begin{tabular}{lp{2.5in}}
\m{1. Episode \rt Event + Reaction} & {\it (An episode might be an event plus a reaction)} \\
\m{2. Reaction \rt Plan + Application} & {\it (A reaction might be a plan plus its application)} \\
\m{3. Application \rt (Preaction)* + Action} & {\it (An application might be zero or more preactions, then an action)} \\
\end{tabular}\end{center} \normalsize
Examples of constraints include that the event
\m{initiates} the reaction, the plan \m{motivates} the
application, and the preactions \m{allow} the action.

Although this style of grammar partially accounts for
our incidents, it has limits as a generative theory.
The grammar is rather unconstrained for generation purposes: 
for example, although stating that ``a reaction is a plan plus its
application'' sounds reasonable, it does not tell the system
exactly what constitutes a plan, or how to generate one.
The situation- and goal-specific nature of plans
suggests a grammar is unlikely to be a good way to
generate this next level of detail in the story;
rather, a planning algorithm would
be needed as a component here, in which case the
bulk of the story generation is being off-loaded
from a grammar to a planner, with the result that
a generative system would end up substantially 
similar to TALE-SPIN. In addition, Rumelhart's
grammar does not distinguish ``good'' and ``bad'' plans
(the `buggy' incidents in Section~\ref{mysystem} would 
be considered well-formed stories by Rumelhart's grammar).
In short, the grammar appears to (partly) account for just the
top level structure of stories, but does not offer information
about constructing the all-important details.

As a top-level account of story structure, Rumelhart's grammar
models events as having clear causal connections (``A happens, 
causing B, thus C does D, hence E,...''). This accounts well
for some stories (eg. Aesop's ``The Man and the Serpent'' 
fable, which he analyzes). However, it does not account
for the introduction of random events (our `happenings'),
not causally related to earlier events. 
Nor (as he describes) does it account well for multi-protagonist
stories where the execution of multiple plans are interleaved,
and possibly interact. Again, in this case, the
strict ordering which a grammar imposes seems inappropriate
for describing the miriad of ways that parallel plan 
executions can interleave. 

\subsection{TAILOR \label{tailor}}

Smith and Witten describe a delightful plan-based story generator,
called TAILOR \cite{tailor}, which has some interesting
similarities with our incident generator described here. 
Like our work (and TAIL-SPIN), stories are generated by
constructing and executing rational plans for characters
in a (toy) world simulation. Again, like our work, a situation
is represented by a set of facts, and actions are represented
by their preconditions and effects on that set of facts.

However, there also some interesting differences in
the details of TAILOR: First, it's planning algorithm
is more like a game-playing algorithm, forward-chaining
from the current situation by trying all possible actions,
evaluating the resulting situations using a scoring function, and then trying
all possible actions from there etc. In this approach,
the evalution function measures how `near' a situation is
to helping a character satisfy his/her goal. The planning
algorithm is thus aware that some actions produce better
results than others, but has no explicit
representation of why. This contrasts with our STRIPS-like planner 
(backward-chaining means-ends analysis), where an action
in the plan is explicitly chosen because it enables a 
subsequent action to be performed. Second, instead of
introducing random `happenings' to liven up a story,
TAILOR includes a second character (besides the `hero'
in the story) who's job is to thwart the hero as
much as possible (although there is no representation
of why this antagonist may want to do this). Each 
character (approximately) 
takes turns during story generation, and thus the
analogy to game-playing is very close -- a story
is in effect a narrative describing the ``game'' the characters
are playing, the hero trying to achieve a goal, while
the antagonist trys to stop him/her at every move. 
This produces some interesting and sometimes amusing results.

\section{Summary}

This working note has provided a brief excursion into
the issue of `story' (episodic narrative) generation, and 
discussed three different approaches (scripts, story
grammars, and a world-simulation/planning approach) in
the context of generating `toy' aviation
incident reports. Of these, the last appears to be most effective for
story generation, although scripts have been shown
useful for story understanding (eg. the systems FRUMP \cite{frump,frump2},
SAM \cite{sam}, and BORIS \cite{boris}).
Common to all three approaches is the need to represent
the goal-related and causal links between events,
highlighting some key knowledge representation requirements
for systems which are to understand episodic narratives
in depth.

\bibliography{ref}

\begin{thebibliography}{}

\bibitem[Cullingford, 1977]{sam}
Cullingford, R.~E. (1977).
\newblock Controlling inference in story understanding.
\newblock In {\em {IJCAI-77}}, page~17.

\bibitem[Dehn, 1981]{dehn}
Dehn, N. (1981).
\newblock Story generation after {TALE-SPIN}.
\newblock In {\em {IJCAI-81}}, volume~1, pages 16--18.

\bibitem[{DeJong}, 1977]{frump2}
{DeJong}, G. (1977).
\newblock Skimming newspaper stories by computer.
\newblock In {\em {IJCAI-77}}, page~16.

\bibitem[{DeJong}, 1979]{frump}
{DeJong}, G. (1979).
\newblock Prediction and substantiation: two processes that comprise
  understanding.
\newblock In {\em {IJCAI-79}}, pages 217--222.

\bibitem[Dyer, 1981]{boris}
Dyer, M.~G. (1981).
\newblock {\$RESTAURANT} revisited, or `lunch with boris'.
\newblock In {\em {IJCAI-81}}, pages 234--236.

\bibitem[Fikes et~al., 1981]{strips}
Fikes, R., Hart, P., and Nilsson, N. (1981).
\newblock Learning and executing generalized robot plans.
\newblock In Webber, B. and Nilsson, N., editors, {\em Readings in AI}, pages
  231--249. Tioga, Palo Alto, CA.

\bibitem[Meehan, 1977]{tale-spin}
Meehan, J.~R. (1977).
\newblock {TALE-SPIN}, an interactive program that writes stories.
\newblock In {\em {IJCAI-77}}, pages 91--98.

\bibitem[Rumelhart, 1975]{rumelhart}
Rumelhart, D.~E. (1975).
\newblock Notes on a schema for stories.
\newblock In Bobrow, D.~G. and Collins, A., editors, {\em Representation and
  Understanding}, pages 211--236. Academic Press, Orlando.

\bibitem[Smith and Witten, 1991]{tailor}
Smith, T.~C. and Witten, I.~H. (1991).
\newblock A planning mechanism for generating story text.
\newblock {\em Literary and Linguistic Computing}, 6(2):119--126.

\bibitem[Wilensky, 1983]{story-grammars}
Wilensky, R. (1983).
\newblock Story grammars versus story points.
\newblock {\em Behavioral and Brain Sciences}, 6:579--623.
\newblock (Paper plus peer commentaries).

\end{thebibliography}

\newpage

\appendix

\section{Prolog Implementation of the Incident Generator}

\small \begin{verbatim}
% File: talespin.pl
% Author: Peter Clark
% Date: Jan 1999
% Purpose: Simple and highly improvised reconstruction of Meehan's TALE-SPIN 
%      story generator, here applied to aviation incident "stories". This
%      reconstruction undoubtedly misses out a lot of Meehan's program, and 
%      also adds in new parts/approaches that Meehan didn't originally use.

talespin :-
        InitialSituation = 
                    [ plocation(passengers1, gate(seattle)),
                      alocation(airplane1, gate(seattle)),
                      flight_path(seattle, chicago),
                      flight_path(chicago, dallas),
                      airplane(airplane1),
                      passengers(passengers1) ],
        InitialGoal = plocation(passengers1, gate(dallas)),
        make_best_plan(InitialGoal, InitialSituation, InitialPlan),
        Prob = 0.3,             % Probability of incident occurring at a particular step
        execute_plan(InitialPlan, InitialSituation, InitialGoal, 
                                Prob, StoryActions, _StorySituations),
        write('Once upon a time...'), nl,
        anglify(StoryActions, StoryText),
        lwrite(StoryText).

% ======================================================================
%               THE STRIPS-LIKE PLANNER
% ======================================================================
% make_plan/3: Simple backward-chaining planner, without a protected goal list.

make_best_plan(Goal, Situation, BestPlan) :-
        findall(Quality-Plan, 
                   ( make_plan(Goal,Situation,Plan),
                     plan_quality(Plan,Quality) ),
                RankedPlans),
        sort(RankedPlans, OrderedPlans),
        last(_-BestPlan, OrderedPlans).

plan_quality(Plan, Quality) :- 
        length(Plan, Length),                                 % lose 10 points per step
        ( member(evacuate(_),Plan) -> Cost = 1 ; Cost = 0 ),  % lose 1 for evacuating
        Quality is 100 - Length*10 - Cost.

% ----------

make_plan(Goal, Situation, Plan) :-
        make_plan(Goal, Situation, [], _FinalSituation, Plan).

make_plan(Goal, Situation, _, Situation, []) :-
        satisfied(Goal, Situation).                    % no cut, as maybe multiple solns
make_plan(Goal, Situation, GoalStack, NewSituation, Actions) :-
        \+ satisfied(Goal, Situation), 
        \+ member(Goal, GoalStack),                           % avoid looping   
        event_definition(action, Action, PCs, Dels, Adds),
        achieves(PCs, Dels, Adds, Goal),
        make_plans(PCs, Situation, [Goal|GoalStack], MidSituation, PreActions),
        apply_effects(Dels, Adds, MidSituation, NewSituation),
        append(PreActions, [Action], Actions).

make_plans([], Situation, _, Situation, []).
make_plans([Goal|Goals], Situation, GoalStack, NewSituation, Actions) :-
        make_plan(Goal, Situation, GoalStack, MidSituation, FirstActions),
        make_plans(Goals, MidSituation, GoalStack, NewSituation, RestActions),
        append(FirstActions, RestActions, Actions).

achieves(_, _, Adds, Goal) :-                           % Goal = effect
        member(Goal, Adds).             
achieves(_, _, Adds, Goal) :-                           % Goal = ramification of effect
        rule(( Goal :- Facts )),
        subset(Facts, Adds).

% ======================================================================
%                       THE SIMULATOR
% This is similar to the planner, *except* it will also throw a spanner
% in the works (ie. a happening), requiring replanning.
% ======================================================================

execute_plan([], FinalSituation, _, _, [], [FinalSituation]) :- 
        !.
execute_plan(_Actions, Sitn, Goal, P, [Happening|NextActions], [Sitn|NextSitns]) :-
        maybe(P),                  % incident happens!! Abandon old Actions and Goal...
        !,
        findall(Happening, 
                   ( event_definition(happening,Happening,PCs,_,_),
                     satisfieds(PCs,Sitn) ),                   % is physically feasible
                Happenings),
        rnd_member(Happening, Happenings),                  % choose a random happening 
        do_event(Happening, Sitn, NewSitn),
        revise_goal(NewSitn, Goal, NewGoal),
        make_best_plan(NewGoal, NewSitn, NewActions),
        execute_plan(NewActions, NewSitn, NewGoal, 0, NextActions, NextSitns).
execute_plan([Action|Actions], Sitn, Goal, P, [Action|NextActions], [Sitn|NextSitns]) :-
        do_event(Action, Sitn, NewSitn),
        execute_plan(Actions, NewSitn, Goal, P, NextActions, NextSitns).

do_event(Event, Situation, NewSituation) :-
        event_definition(_, Event, PCs, Dels, Adds),
        apply_effects(Dels, Adds, Situation, NewSituation).

% Do the deletes and adds as appropriate
apply_effects(Dels, Adds, Situation, NewSituation) :-
        removes(Dels, Situation, MidSituation),
        append(Adds, MidSituation, NewSituation).

% ======================================================================
%               OTHER UTILITIES
% ======================================================================

satisfieds([], _).
satisfieds([F|Fs], S) :- 
        satisfied(F, S), 
        satisfieds(Fs, S).

satisfied(Fact, Situation) :- 
        member(Fact, Situation).                        
satisfied(Fact, Situation) :-
        rule((Fact :- Facts)),                    % Fact is a ramification of the world
        satisfieds(Facts, Situation).

% ---------- writing...

lwrite([]).
lwrite([X|Xs]) :- write('       '), lwrite2(X), nl, lwrite(Xs).

lwrite2([]).
lwrite2([BitX|BitXs]) :- !, write(BitX), lwrite2(BitXs).
lwrite2(X) :- write(X).

anglify([], []).
anglify([Event|Events], [English|Englishs]) :-
        event_english(Event, English),
        anglify(Events, Englishs).

% ======================================================================
%               GENERAL UTILITIES
% ======================================================================

removes([], L, L).
removes([R|Rs], L, NewL) :-
        remove(R, L, MidL),
        removes(Rs, MidL, NewL).

remove(A, [A|B], B).
remove(A, [C|B], [C|NewB]) :-
        remove(A, B, NewB).

member(X, [X|_]).
member(X, [_|Y]) :- member(X, Y).

memberchk(X, Y) :- member(X, Y), !.

subset([], _).
subset([X|Xs], Ys) :- remove(X, Ys, RestYs), subset(Xs, RestYs).

nmember(Elem, List, N) :-
        nmember(Elem, List, 1, N).

nmember(Elem, [Elem|_], N, N).
nmember(Elem, [_|List], NSoFar, N) :-
        NewN is NSoFar + 1,
        nmember(Elem, List, NewN, N).

last(X, [X]).
last(A, [_,C|D]) :-
        last(A, [C|D]).

% ---------- Randomization utilities 

:- dynamic lastrnd/1.
lastrnd(0).

maybe(P) :- random(R), R < P, !.                        % succeed with probability P

random(R) :- 
        lastrnd(N), rnd_number(N,R), retract(lastrnd(N)), NewN is N + 1, 
        ( NewN >= 20 -> assert(lastrnd(0)) ; assert(lastrnd(NewN)) ).
        
rnd_member(X, Xs) :-
        length(Xs, L),
        random(R),
        N is integer(R*L) + 1,
        nmember(X, Xs, N), !.

rnd_number( 0,0.174232). rnd_number( 1,0.186011). rnd_number( 2,0.951800). 
rnd_number( 3,0.363587). rnd_number( 4,0.108449). rnd_number( 5,0.848878). 
rnd_number( 6,0.309133). rnd_number( 7,0.230964). rnd_number( 8,0.639224). 
rnd_number( 9,0.686739). rnd_number(10,0.781066). rnd_number(11,0.983691). 
rnd_number(12,0.704568). rnd_number(13,0.636376). rnd_number(14,0.881027). 
rnd_number(15,0.194111). rnd_number(16,0.449212). rnd_number(17,0.110336). 
rnd_number(18,0.572139). rnd_number(19,0.149503). 
\end{verbatim} \normalsize

\newpage

\section{Prolog KB for Aviation Incidents}

\small \begin{verbatim}
% File: talespin-kb.pl
% Author: Peter Clark
% Date: Jan 1999
% Purpose: Knowledge Base about Aviation incidents for talespin.pl

% ======================================================================
%               THE FLIGHT INCIDENT KNOWLEDGE BASE
% ======================================================================

event_definition(Type, Event, PCs, Adds, Dels) :- ed(Type, Event, PCs, Adds, Dels, _).
event_english(Event, English) :- ed(_, Event, _, _, _, English).

% ---------- Routine actions... ----------

ed(action, load(Passengers,Airplane), 
        /*pcs*/ [plocation(Passengers,gate(Airport)),alocation(Airplane,gate(Airport))],
        /*del*/ [plocation(Passengers,gate(Airport))],
        /*add*/ [contains(Airplane,Passengers)],
        /*txt*/ 'The passengers boarded the plane.' ).

ed(action, taxi_to_runway(Airplane),
        /*pcs*/ [alocation(Airplane,gate(Airport))],
        /*del*/ [alocation(Airplane,gate(Airport))],
        /*add*/ [alocation(Airplane,runway(Airport))],
        /*txt*/ 'The plane taxiied to the runway.' ).

ed(action, take_off(Airplane,Airport),
        /*pcs*/ [alocation(Airplane,runway(Airport))],
        /*del*/ [alocation(Airplane,runway(Airport))],
        /*add*/ [alocation(Airplane,near(Airport))],
        /*txt*/ ['The plane took off from ',Airport,'.']).

ed(action, cruise(Airplane,Airport1,Airport2),
        /*pcs*/ [flight_path(Airport1,Airport2),alocation(Airplane,near(Airport1))],
        /*del*/ [alocation(Airplane,near(Airport1))],
        /*add*/ [alocation(Airplane,near(Airport2))],
        /*txt*/ ['The plane cruised towards ',Airport2,'.']).

ed(action, land(Airplane,Airport2),
        /*pcs*/ [alocation(Airplane,near(Airport2))],
        /*del*/ [alocation(Airplane,near(Airport2))],
        /*add*/ [alocation(Airplane,runway(Airport2))],
        /*txt*/ ['The plane landed at ',Airport2,'.']).

ed(action, taxi_to_gate(Airplane),
        /*pcs*/ [alocation(Airplane,runway(Airport))],
        /*del*/ [alocation(Airplane,runway(Airport))],
        /*add*/ [alocation(Airplane,gate(Airport))],
        /*txt*/ 'The plane taxiied to the gate.' ).

ed(action, unload(Passengers,Airplane), 
        /*pcs*/ [contains(Airplane,Passengers),alocation(Airplane,gate(Airport))],
        /*del*/ [contains(Airplane,Passengers)],
        /*add*/ [plocation(Passengers,gate(Airport))],
        /*txt*/ 'The passengers disembarked.' ).

% ---------- Emergency actions... ----------

ed(action, evacuate(Airplane),
        /*pcs*/ [a_on_ground(Airplane),alocation(Airplane,Loc),contains(Airplane,Passengers)],
        /*del*/ [contains(Airplane,Passengers)],
        /*add*/ [plocation(Passengers,Loc)],
        /*txt*/ 'The passengers were evacuated from the plane.' ).

ed(action, emergency_landing(Airplane),
        /*pcs*/ [alocation(Airplane,near(Airport2))],
        /*del*/ [alocation(Airplane,near(Airport2))],
        /*add*/ [alocation(Airplane,on_ground_near(Airport2))],
        /*txt*/ ['The pilot made an emergency landing near ',Airport2,'.']).

ed(action, medical_help(Passengers),
        /*pcs*/ [plocation(Passengers, gate(_))],               % any gate
        /*del*/ [],
        /*add*/ [medical_help(Passengers)],
        /*txt*/ 'Medical help was provided.' ).

% ---------- Possible happenings... ----------

ed(happening, fire(engine),
        /*pcs*/ [],                                     % can happen anywhere
        /*del*/ [],
        /*add*/ [on_fire(engine)], 
        /*txt*/ 'The engine caught fire.' ).

% ---------- Possible happenings... ----------

ed(happening, ill_passenger,
        /*pcs*/ [contains(Airplane,Passengers),passengers(Passengers),airplane(Airplane)],
        /*del*/ [],
        /*add*/ [ill_passenger],
        /*txt*/ 'A passenger became very ill.' ).

% Ramifications of facts about the world...
rule(( a_on_ground(Airplane) :- [alocation(Airplane,gate(_))] )).
rule(( a_on_ground(Airplane) :- [alocation(Airplane,runway(_))] )).
rule(( a_on_ground(Airplane) :- [alocation(Airplane,on_ground_near(_))] )).
rule(( p_on_ground(Passengers) :- [plocation(Passengers,gate(_))] )).
rule(( p_on_ground(Passengers) :- [plocation(Passengers,runway(_))] )).
rule(( p_on_ground(Passengers) :- [plocation(Passengers,on_ground_near(_))] )).

% Rules for revising the goal
revise_goal(Situation, plocation(Passengers,_), Goal) :- % If the engine's on fire, 
        memberchk(on_fire(engine), Situation), !,        % get to the ground asap!
        Goal = p_on_ground(Passengers).
revise_goal(Situation, plocation(Passengers,_), Goal) :- % If a passenger's ill, 
        memberchk(ill_passenger, Situation), !,          % get to a gate somewhere.
        Goal = medical_help(Passengers).
revise_goal(_Situation, Goal, Goal).
\end{verbatim} \normalsize

\end{document}